\documentclass[11pt,a4paper]{article}
\usepackage{times,latexsym}
\usepackage{url}
\usepackage[T1]{fontenc}

\usepackage[acceptedwithA]{tacl2021v1}

\usepackage{xspace,mfirstuc,tabulary}

\newif\iftaclinstructions
\taclinstructionsfalse 
\iftaclinstructions
\renewcommand{\confidential}{}
\renewcommand{\anonsubtext}{(No author info supplied here, for consistency with
TACL-submission anonymization requirements)}
\newcommand{\instr}
\fi

\iftaclpubformat 

\else

\fi

\usepackage{enumitem}
\setitemize{noitemsep,topsep=0pt,parsep=0pt,partopsep=0pt}
\definecolor{cite}{rgb}{0.6,0.6,1.0}

\definecolor{todo}{rgb}{1,0.5,0}

\usepackage{booktabs}
\usepackage{graphicx}
\usepackage{float}
\usepackage{tabularx}
\usepackage{soul}
\usepackage{subcaption}
\usepackage{multicol}
\usepackage{multirow}
\usepackage{amsfonts}
\usepackage{tabularx}
\usepackage{makecell}

\def\approachname{\textsc{Citention}}
\def\benchmarkname{\textsc{CiteControl}}
\def\reck{R$k$}
\def\reckf{R$k^\mathrm{f}$}

\def\repliqa{RepliQA}
\def\boolq{BoolQ}
\def\boolqm{BoolQ-M}
\def\musique{MuSiQue}
\def\neoqa{NeoQA}
\def\qasper{QASPER}
\def\govreport{GovReport}
\def\gen{\texttt{GEN}}
\def\icr{\texttt{ICR}}

\def\at{\texttt{AT2}}
\def\qr{\texttt{QR}}
\def\drag{\texttt{DRAG}}
\def\bmr{\texttt{BM25}}
\def\comba{\texttt{COMB-A}}
\def\combr{\texttt{COMB-R}}
\def\comb{\texttt{COMB}}
\def\single{\texttt{single}}
\def\concatenation{\texttt{multi-hop}}
\def\intersection{\texttt{intersection}}
\def\implicit{\texttt{implicit}}
\def\explicit{\texttt{explicit}}
\usepackage{bm}
\usepackage{amsmath}
\usepackage{hyperref}
\usepackage[most]{tcolorbox}

\usepackage{float}

\title{Citation Failure: Definition, Analysis and Efficient Mitigation}

\author{
  Jan Buchmann 
  \and
  Iryna Gurevych
  \\
  \ \\
  Ubiquitous Knowledge Processing Lab (UKP Lab)
  \\
  Department of Computer Science and Hessian Center for AI (hessian.AI)
  \\
  Technical University of Darmstadt 
  \\
  \url{www.ukp.tu-darmstadt.de}
  \\
  \ 
}

\date{}

\begin{document}
\maketitle
\begin{abstract}

Citations from LLM-based RAG systems are supposed to simplify response verification. However, this goal is undermined in cases of \textit{citation failure}, where a model generates a helpful response, but fails to generate citations to complete evidence. In contrast to previous work, we propose to disentangle this from \textit{response failure}, where the response itself is flawed, and citing complete evidence is impossible. To address citation failure, this work follows a two-step approach: (1) We study when citation failure occurs and (2) how it can be mitigated efficiently. For step 1, we extend prior work by investigating how the relation between response and evidence affects citation quality. We introduce \benchmarkname{}, a benchmark that systematically varies this relation to enable the analysis of failure modes. Experiments show that failures increase with relational complexity and suggest that combining citation methods could improve performance, motivating step 2. To study the efficient improvement of LLM citation, we propose \approachname{}, a framework integrating generative, attention-based, and retrieval-based methods. Results demonstrate substantial citation improvements on \benchmarkname{} and in transfer settings. We make our data and code publicly available.\footnote{\url{https://github.com/UKPLab/tacl2026-citation-failure}}

\end{abstract}

\section{Introduction}
\label{sec:introduction}

Citations for LLM-generated responses can improve usefulness and accountability in RAG \cite{lewis2020rag} by enabling users to verify responses through evidence. However, this advantage weakens when a model produces a valid response without sufficient cited evidence: users must then either search additional sources or discard an otherwise helpful answer. To address this important issue, we adopt a two-step approach: We (1) analyze when models generate insufficient citations, and (2) investigate efficient mitigation strategies.


\paragraph{Step 1: Analyzing insufficient citations} 

Approaches to obtaining citations for LLM-generated responses (also known as attributions; \citealt{rashkin2023attribution}) operate in three distinct paradigms: \\
(1) In \textit{generative citation}, models generate responses and citations in a single step \cite{huang-etal-2024-learning, droste2026sui1groundedverifiablelongform}.
(2) \textit{Attribution-first} approaches prompt LLMs to select evidence pieces before generation, thereby enforcing a strict mapping between evidence and response \cite{slobodkin-etal-2024-attribute, wright-etal-2025-unstructured}. (3) \textit{Post-hoc}-approaches use LLMs \cite{balasubramanian2025decompositionenhancedtrainingposthocattributions, hirsch2025laquerlocalizedattributionqueries} or retrievers \cite{ramu-etal-2024-enhancing, sancheti-etal-2024-post} to retrieve citations for responses generated by a different model. LLM-based attribution-first and post-hoc approaches suffer from high complexity and long inference times, as they require multiple LLM calls. Therefore, we focus on generative citation, which is simpler and faster, and use retrieval-based post-hoc approaches as efficient baselines.

\begin{figure}
    \centering
    \includegraphics[scale=0.95]{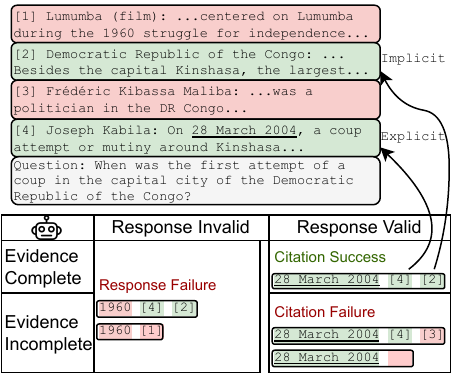}
    \caption{Citation Example:\footnotemark An LLM receives multiple documents and a question. The confusion matrix shows the possible outcomes for generated response and evidence. The response-evidence relation has reasoning type \concatenation{}. It is \explicit{} for the response and [4], and \implicit{} for the response and [2].}
    \label{fig:eyecatcher}
\end{figure}

Prior work that studies generative citation does not differentiate between \textit{citation failure} and \textit{response failure} (e.g. \citealt{gao2023alce, zhang2024longciteenablingllmsgenerate, wu2025reflongbenchmarkinglongcontextreferencing}). We believe that this is an important gap. To illustrate this distinction, consider a question about the time of a coup in the capital of the DR Congo and source documents from a web search (Fig.~\ref{fig:eyecatcher}). Multi-hop reasoning is required: Document [2] states that Kinshasa is the capital, while [4] reports a coup there on 28 March 2004. In this situation, an LLM may exhibit: (1) \textit{Response failure}: where the generated response is invalid, i.e. not supported by any combination of source documents (e.g. "1960"), and any cited evidence cannot support it. (2) \textit{Citation success}: where the response is valid ("28 March 2004") and the evidence is complete ("[2] [4]"). (3) \textit{Citation failure}: where the response is valid, but the evidence is incomplete (e.g. by citing [3] instead of [2]). 

Most work analyzing insufficient citations focuses on source document properties and composition (e.g., \citealt{wu2025reflongbenchmarkinglongcontextreferencing, sorodoc2025garagebenchmarkgroundingannotations}). In LLM-based post-hoc citation, the relationship between evidence and statements—such as varying reasoning complexity \cite{zhu-etal-2025-trove, balasubramanian2025decompositionenhancedtrainingposthocattributions}—strongly affects citation performance. However, its impact on generative citation remains largely unexplored and requires further study.

Two aspects of existing datasets and benchmarks make them unsuitable for this analysis: First, they do not allow rigorous verification of response correctness, making it hard to separate response from citation failure. Second, they rely on LLM-based evaluators \cite{tang-etal-2024-minicheck, honovich2022true}, whose accuracy can drop to $\sim$50\% in complex cases \cite{hu2025llmsevaluatecomplexattribution}. As a result, a dedicated benchmark and analysis of how the response–evidence relationship affects citation failure is needed.

To address these gaps, we ask:


\textbf{RQ1: How does the response–evidence relation affect citation failure?} Building on prior work in citation evaluation \cite{hu2025llmsevaluatecomplexattribution} and intertextuality \cite{kuznetsov2022intertext}, we define \textit{reasoning type} and \textit{overtness} as core properties of this relation and introduce \benchmarkname{}, a benchmark that systematically varies them. All instances include verifiable answers and known evidence, enabling us to control for response failure and avoid reliance on error-prone citation evaluators.

Experiments on \benchmarkname{} with generative citation from 18 LLMs across five families, along with retrieval-based post-hoc baselines, validate the need to separate citation from response failure: citation performance is consistently higher for correctly answered instances. The results also reveal distinct failure modes: smaller models struggle even with simple 1-to-1 relations, while all models face difficulties in multi-document reasoning and tend to under-generate citations. Comparing generative and retrieval-based methods shows that different approaches suit different relation types, indicating that combining them can improve performance.

\footnotetext{\label{fn:robot_flaticon}Robot icon created by Good Ware - \href{https://www.flaticon.com/free-icons/robot}{Flaticon}}

\paragraph{Step 2: Mitigating citation failure efficiently}
Most approaches to improving generative citation rely on fine-tuning (e.g., \citealt{huang-etal-2024-learning, zhang2024longciteenablingllmsgenerate, li-etal-2024-improving-attributed}), which involves complex data collection and training setups, risks catastrophic forgetting \cite{biderman2024lora, dou-etal-2024-loramoe}, and demands substantial resources. Similarly, attribution-first and post-hoc methods often require multiple additional LLM calls, making them costly. In contrast, retrieval-based approaches \cite{ramu-etal-2024-enhancing, sancheti-etal-2024-post} are efficient but limited by their small parameter capacity. This highlights the need for methods that are both efficient and powerful in mitigating citation failure.

In related settings, several works have successfully used attention values for reranking \cite{zhang2025queryfocusedretrievalheadsimprove, chen2025attention} and retrieving parts of the context that most influenced an LLM's output \cite{cohenwang2025learningattributeattention}. This efficiently exploits the capabilities of LLMs, as attention values are available ``for free'' during generation,\footnote{In practice, attention implementations such as Flash Attention \cite{dao2023flashattention2} do not give access to attention values, making an additional forward pass the most time-efficient implementation to date \cite{cohenwang2025learningattributeattention}.} but the potential of these approaches to mitigate citation failure has not been studied. Therefore, we aim to answer the research question:

\textbf{RQ2: How can attention values be used to efficiently mitigate LLM citation failure?} We contribute \approachname{}, a framework designed for studying this problem that integrates generative, attention-based, and retrieval-based citation. We perform experiments with \approachname{} across 6 LLMs (1.7–70B parameters) on \benchmarkname{} and two transfer datasets.

Our key findings are: First, attention-based citation is effective for extractive and abstractive responses (at least $+$10\% average relative improvement over generative citation on transfer datasets), but partially fails with complex reasoning and short abstractive responses. Second, combining generation-, attention-, and retrieval-based citation reliably improves over generative citation, including complex reasoning cases (at least $+$5\% relative improvement averaged over all datasets). Third, masking reasoning tokens during attention computation improves attention-based citation. These findings provide clear guidance for future research on efficient LLM citation.

\paragraph{} In summary, we make two main contributions: (1) \benchmarkname{}, a novel benchmark for LLM citation (§\ref{sec:benchmark}), on which experiments reveal that there is ample room for improvement in cases with complex statement-evidence relations (§\ref{sec:citation_analysis}). (2) \approachname{}, a framework for studying the efficient mitigation of citation failure with attention-based and retrieval-based methods (§\ref{sec:approach}), and experiments on \benchmarkname{} and in a transfer setting showing that failures from generation-based citation can be mitigated effectively (§\ref{sec:attention}).



\section{Related Work}
\label{sec:related_work}

\paragraph{Citations} The goal of citations is to provide \textit{corroborative} attribution, i.e. retrieving evidence $E$ for a statement $s$, such that ``according to $E$, $s$'' is true \cite{rashkin2023attribution}. \textit{Contributive} attribution is a related, but distinct paradigm: Here, the goal is to estimate the effect of parts of the context on the LLM output, often measured by changes in output probability when removing these parts \cite{cohenwang2024contextcite, ribeiro2016whyshouldi}.

\paragraph{Analyzing Generative LLM Citation} A range of datasets \cite{malaviya-etal-2024-expertqa, kamalloo2023hagridhumanllmcollaborativedataset, golany-etal-2024-efficient} and benchmarks \cite{gao2023alce, liu-etal-2023-evaluating} for comparing LLMs in their citation abilities have been proposed. Existing analyses focus on the properties of the source documents such as their number and combination \cite{koo-etal-2024-large, tang2024lciteevallongcontextmodelstruly, buchmann-etal-2024-attribute, wu2025reflongbenchmarkinglongcontextreferencing}, authorship information \cite{abolghasemi2024evaluationattributionbiasretrievalaugmented}, and time-dependence and semantic properties of their contents \cite{sorodoc2025garagebenchmarkgroundingannotations} as factors influencing citation failure. While the effect of the response-evidence relation has been studied in post-hoc citation \cite{zhu-etal-2025-trove, balasubramanian2025decompositionenhancedtrainingposthocattributions}, this has not been studied for generative citation.

\paragraph{Improving Citations} To improve citations for LLM responses, (1) training-based approaches collect data and design training regimes to improve generative citation capabilities \cite{huang-etal-2024-learning, penzkofer-baumann-2024-evaluating, zhang2024longciteenablingllmsgenerate, li-etal-2024-improving-attributed} (2) LLM-based attribution-first \cite{slobodkin-etal-2024-attribute, wright-etal-2025-unstructured} and post-hoc approaches split attribution across multiple LLM calls \cite{hirsch2025laquerlocalizedattributionqueries, qian-etal-2024-grounding}. (3) Contributive-attribution based methods ablate context across multiple forward passes to isolate relevant sources \cite{chuang2025selfciteselfsupervisedalignmentcontext, qi-etal-2024-model}. (4) Retrieval-based post-hoc approaches use sparse or dense retrievers post-generation \cite{sancheti-etal-2024-post, ramu-etal-2024-enhancing}. Categories (1–3) are resource-intensive at training (1) or inference (2-3), while (4) is constrained by the retriever’s capacity, typically smaller than the LLM’s.

\paragraph{Using LLM Internals for Efficient Retrieval} Several works have proposed directly using LLM internals such as attention values or hidden states on related problems such as reranking \cite{zhang2025queryfocusedretrievalheadsimprove, chen2025attention} and contributive attribution \cite{cohenwang2025learningattributeattention, phukan-etal-2024-peering, ding2024attentiondependencyparsingaugmentation}. This exploits LLM capabilities efficiently, as no training of the LLM itself or additional LLM calls are needed. \citet{hirsch2025laquerlocalizedattributionqueries} recently showed that the hidden-states-based method from \citet{phukan-etal-2024-peering} shows mediocre performance in a fine-grained setting, so we do not consider it here. To our knowledge, the use of attention-based methods for citation has not been investigated. 

\paragraph{} We make important contributions to the described research areas: In analyzing LLM citation, we are the first to provide response-failure-aware methodology and experiments investigating the effect of the response-evidence relation on citation failure. The results inform our research on improving LLM citation, where we investigate the potential of attention-based methods for corroborative citation and its combination with generation-based and retrieval-based citation for the first time.


\section{\benchmarkname{}}
\label{sec:benchmark}

\begin{table*}
    \centering
    \small
    \begin{tabular}{l|ccccccc}
\toprule
 & \# train/dev/test & $|S|$ & $|s|$ & $|r|$ & $|E|$ & Reasoning & Overtness \\
\midrule
RepliQA & 6084/1933/2087 & 20.0 & 84.3 & 5.1 & 1.0 & \single{} & \explicit{} \\
BoolQ-M & 3394/849/2173 & 20.0 & 96.4 & 1.0 & 1.0 & \single{} & \implicit{} \\
MuSiQue & 15950/3988/2417 & 20.0 & 82.4 & 2.3 & 2.4 & \texttt{multi-hop} & \texttt{exp} / \texttt{imp} \\
NeoQA & 0/264/1157 & 19.7 & 283.4 & 6.3 & 1.9 & \texttt{multi-hop} / \texttt{intersec.} & \texttt{exp} / \texttt{imp} \\
\bottomrule
\end{tabular}
    \caption{The datasets in \benchmarkname{}. $|S|$ / $|E|$: Number of source / evidence documents per instance. $|s|$ / $|r|$: Number of words per source document / response.}
    \label{tab:datasets}
\end{table*}

To study how the response-evidence relation affects citation, we introduce \benchmarkname{}, a framework for evaluating and analyzing LLM citation. Unlike prior benchmarks \cite{gao2023alce, tang2024lciteevallongcontextmodelstruly, buchmann-etal-2024-attribute}, it separates response failure from citation failure, and avoids reliance on error-prone attribution models \cite{hu2025llmsevaluatecomplexattribution}. We first formalize the citation task (§\ref{sec:formalization}), then detail how we vary response–evidence relations (§\ref{sec:benchmark/design}), followed by datasets (§\ref{sec:benchmark/datasets}) and evaluation (§\ref{sec:benchmark/evaluation}).

\subsection{Task Formalization}
\label{sec:formalization}

An instance in \benchmarkname{} consists of an instruction $q$ (e.g. a question) and a set of source documents $S=\{s_1,...,s_{|S|}\}$ (these could come from a previous retrieval step). The task is to generate a \textit{response} $r$ based on $S$ (e.g. an answer), and to cite corroborative \textit{evidence} $E \subset S$ (see §\ref{sec:related_work}, \citealt{rashkin2023attribution}). The evidence should enable response verification, so models need to cite relevant source documents even if the contained information is present in their parametric knowledge. 

\subsection{Varying the Relation between Response and Evidence}
\label{sec:benchmark/design}

We build on related work to define two key properties of the response-evidence relation and study their effect on LLM citation:

\paragraph{Reasoning Type} Following \citet{hu2025llmsevaluatecomplexattribution}, we distinguish the types of reasoning required to infer the response from the evidence, omitting ``union'' due to lack of suitable data: 
\begin{itemize}
    \item \single{}: Reasoning over a single evidence document. 
    \item \concatenation{}:\footnote{Named ``concatenation'' in \citet{hu2025llmsevaluatecomplexattribution}} Reasoning over a chain of facts, the final fact being the correct response (as in Fig.~\ref{fig:eyecatcher}).
    \item \intersection{}: Combining multiple facts to a new ``fact'' (e.g. computing the time between two events).
\end{itemize}

\paragraph{Overtness} \citet{hu2025llmsevaluatecomplexattribution} assume verbatim extraction from evidence documents, and do not differentiate their analysis between individual evidence documents. This misses the \textit{overtness} of the response-evidence relation, which is recognized by more general research on intertextuality \cite{kuznetsov2022intertext}, and distinguishes:
\begin{itemize}
    \item \implicit{}: the response does not appear verbatim, but the evidence document is relevant (e.g document [2] in Fig.~\ref{fig:eyecatcher})
    \item \explicit{}: the response appears verbatim in the evidence document (e.g. [4] in Fig.~\ref{fig:eyecatcher})
\end{itemize}



\begin{figure}
    \centering
    \includegraphics[width=\columnwidth]{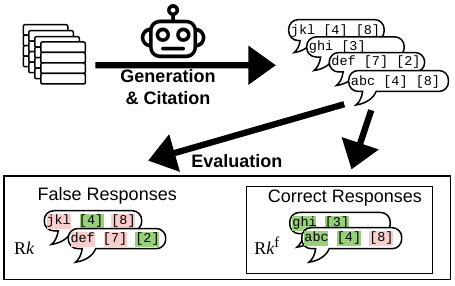}
    \caption{Evaluation strategy on \benchmarkname{}:\textsuperscript{\ref{fn:robot_flaticon}} For \reck{}, all predictions are evaluated for evidence recall @$k$, while for \reckf{}, the subset of predictions with correct responses are evaluated.}
    \label{fig:eval_strategy}
\end{figure}

\subsection{Datasets}
\label{sec:benchmark/datasets}

Datasets in \benchmarkname{} must (1) provide known reasoning types and overtness (§\ref{sec:benchmark/design}), (2) include verifiable responses to separate response from citation failure, and (3) specify complete ground-truth evidence to avoid reliance on error-prone evaluation models (§\ref{sec:introduction}). Based on these criteria, we select four datasets. See Tables \ref{tab:datasets} for a dataset overview and \ref{tab:format_explanations} for examples.

\textbf{\repliqa} \cite{montciro-nips-2024-repliqa} and \textbf{\boolqm{}} consist of tuples of a context paragraph, question and answer. While \repliqa{} answers are extractive, \boolqm{} answers are ``yes'' or ``no''. We created \boolqm{} to control for data contamination in \boolq{} \cite{clark-etal-2019-boolq} by replacing entities with fictional entities using  GPT-OSS-120B\footnote{\url{https://huggingface.co/openai/gpt-oss-120b}}. \textbf{\musique{}} \cite{trivedi-etal-2022-musique} is a dataset of 2- to 4-hop questions and answers combined with 2 to 4 evidence paragraphs and 16 to 18 distractor paragraphs from Wikipedia. \textbf{\neoqa{}} \cite{glockner2025neoqaevidencebasedquestionanswering} is a dataset of time-span and 2-hop questions on synthetic news articles. For time-span questions, models are given two events and need to compute the time span between them. For \repliqa{}, \boolqm{} and \neoqa{}, we combine evidence documents with distractors to obtain 20 source documents per instance. For details on data processing, see §\ref{sec:replication/data_processing}.

The reasoning type in \repliqa{} and \boolqm{} is \single{}. All \musique{} instances and \neoqa{} 2-hop instances are \concatenation{}, while \neoqa{} time-span instances are \intersection{}.

The overtness of the response-evidence relation can be seen in the ROUGE-1 scores \cite{lin-2004-rouge} between response and evidence shown in Table~\ref{tab:rouge_scores}. It is \explicit{} for \repliqa{} and between the response and the evidence document that contains the final answer in \musique{} and \neoqa{} \concatenation{} instances. It is \implicit{} for evidence documents upstream in the \concatenation{} reasoning chain, as well as for \boolqm{} and \neoqa{} \intersection{} instances. 

\subsection{Evaluation}
\label{sec:benchmark/evaluation}

We propose \textit{filtered} recall @$k$ (\reckf{}) as our primary evaluation metric (Fig.~\ref{fig:eval_strategy}), which improves over existing recall-focused evaluation of citations \cite{gao2023alce, buchmann-etal-2024-attribute} by disentangling response failure from citation failure: For a specific model, we only evaluate the subset of instances that were answered correctly. \textit{@}$k$ means that for any instance, we only evaluate the first $k$ citations to avoid rewarding over-generation of evidence. For reference, we report recall @$k$ on all instances (\reck{}) and the proportion of correctly answered instances ($\frac{|\mathcal{R}^\mathrm{f}|}{|\mathcal{R}|}$). For details, see §\ref{sec:replication/filtered_evaluation}. 


\section{How Does the Relation Between Response Statement and Evidence Affect LLM Citation?}
\label{sec:citation_analysis}

In this section, we use \benchmarkname{} to analyze the effect of the response-evidence relation on citation performance. We describe experimental details in §\ref{sec:citation_analysis/setup} and results in §\ref{sec:citation_analysis/results}.

\subsection{Experimental Setup}
\label{sec:citation_analysis/setup}

\paragraph{Prompts} We instruct models to cite by appending document indices to response statements (Fig.~\ref{fig:eyecatcher}) with 3-shot prompts. For details and examples, see §\ref{sec:replication/prompts}.

\paragraph{Retrieval-based oracle baselines} We include \bmr{}, a lexical-matching based retriever \cite{robertson2009bm25}, and Dragon (\drag{}, \citealt{lin-etal-2023-train}), a dense retriever, which have been shown to perform well on a recent benchmark \cite{buchmann-etal-2024-attribute}. We use the concatenation of question and ground truth answer as the query (for details see §\ref{sec:replication/method_details/retrieval}).

\subsection{Results and Discussion}
\label{sec:citation_analysis/results}

\begin{table*}[!t]
    \centering
    \small
    \begin{tabular}{l|ccc|ccc|ccc|ccc}
\toprule
 & \multicolumn{3}{c}{\textbf{\repliqa}} & \multicolumn{3}{c}{\textbf{\boolqm}} & \multicolumn{3}{c}{\textbf{\musique}} & \multicolumn{3}{c}{\textbf{\neoqa}} \\
 & \textbf{R$\bm k$} & $\bm{\frac{|\mathcal{R}^\mathrm{f}|}{|\mathcal{R}|}}$ & \textbf{R$\bm{k^\mathrm{f}}$}  & \textbf{R$\bm k$} & $\bm{\frac{|\mathcal{R}^\mathrm{f}|}{|\mathcal{R}|}}$ & \textbf{R$\bm{k^\mathrm{f}}$}  & \textbf{R$\bm k$} & $\bm{\frac{|\mathcal{R}^\mathrm{f}|}{|\mathcal{R}|}}$ & \textbf{R$\bm{k^\mathrm{f}}$} & \textbf{R$\bm k$} & $\bm{\frac{|\mathcal{R}^\mathrm{f}|}{|\mathcal{R}|}}$ & \textbf{R$\bm{k^\mathrm{f}}$}  \\
 \midrule
& \multicolumn{12}{c}{No Reasoning} \\
\midrule
Ministr-8B & 96.9 & 73.2 & 98.3 & 99.7 & 81.9 & 99.8 & 31.3 & 26.7 & 44.0 & 51.0 & 47.6 & 52.5 \\
Mistr-S-3.2-24B & 99.2 & 85.2 & 99.9 & 100 & 87.3 & 99.9 & 61.5 & 50.0 & 70.1 & 75.6 & 71.7 & 73.2 \\
Llama-3.2-1B & 39.1 & 63.1 & 41.4 & 43.7 & 58.4 & 44.6 & 10.0 & 11.5 & 14.1 & 8.0 & 19.1 & 8.6 \\
Llama-3.2-3B & 96.0 & 73.0 & 97.1 & 98.5 & 70.8 & 98.6 & 39.1 & 24.7 & 46.0 & 47.9 & 40.5 & 48.6 \\
Llama-3.1-8B & 98.9 & 78.2 & 99.6 & 99.0 & 80.1 & 99.3 & 44.7 & 36.4 & 54.3 & 63.6 & 55.4 & 61.6 \\
Llama-3.3-70B & 97.0 & 79.4 & 98.2 & 98.4 & 78.4 & 98.8 & 41.1 & 29.6 & 49.1 & 61.5 & 52.8 & 61.4 \\
Phi-4-Mini & 93.7 & 68.9 & 95.4 & 92.8 & 63.2 & 97.6 & 41.2 & 21.8 & 53.5 & 61.8 & 42.5 & 63.0 \\
Phi-4 & 97.8 & 80.6 & 99.2 & 98.5 & 79.0 & 99.4 & 64.7 & 36.3 & 75.4 & 73.7 & 49.4 & 72.2 \\
Oracle-BM25 & 98.5 & 100 & 98.5 & 68.3 & 100 & 68.3 & 48.7 & 100 & 48.7 & 70.8 & 100 & 70.8 \\
Oracle-DRAG & 99.4 & 100 & 99.4 & 100 & 100 & 100 & 70.4 & 100 & 70.4 & 68.0 & 100 & 68.0 \\
\midrule
& \multicolumn{12}{c}{Reasoning} \\
\midrule
Qwen3-0.6B & 75.9 & 14.2 & 83.5 & 88.2 & 70.0 & 89.9 & 30.5 & 1.1 & 50.0 & 27.6 & 37.0 & 35.7 \\
Qwen3-1.7B & 91.5 & 70.5 & 94.5 & 95.5 & 81.4 & 96.1 & 31.8 & 29.7 & 40.9 & 45.5 & 48.3 & 53.0 \\
Qwen3-4B & 97.0 & 71.2 & 99.0 & 99.2 & 87.5 & 99.6 & 44.0 & 34.7 & 63.2 & 70.1 & 67.2 & 75.2 \\
Qwen3-8B & 96.9 & 80.2 & 98.6 & 98.8 & 87.0 & 99.8 & 54.9 & 50.4 & 68.0 & 74.5 & 77.4 & 80.8 \\
Qwen3-14B & 98.1 & 81.2 & 99.6 & 99.0 & 88.4 & 99.7 & 57.8 & 51.3 & 69.7 & 75.4 & 81.4 & 79.7 \\
Qwen3-32B & 98.7 & 80.5 & 99.8 & 99.8 & 90.0 & 99.9 & 62.5 & 54.4 & 76.1 & 75.9 & 84.8 & 81.4 \\
GPT-OSS-20B & 96.1 & 77.0 & 97.6 & 99.3 & 84.1 & 99.6 & 51.1 & 48.3 & 63.3 & 77.1 & 57.2 & 82.2 \\
GPT-OSS-120B & 98.6 & 78.2 & 99.6 & 99.9 & 90.0 & 100 & 66.2 & 63.8 & 74.2 & 88.5 & 86.0 & 91.1 \\
GPT-5-Mini & 99.0 & 81.1 & 99.6 & 99.9 & 89.2 & 99.9 & 73.2 & 64.5 & 80.5 & 89.0 & 88.1 & 91.0 \\
GPT-5.2 & 99.5 & 82.8 & 100 & 100 & 90.6 & 100 & 80.2 & 68.6 & 87.0 & 92.8 & 96.1 & 93.8 \\
\bottomrule
\end{tabular}
    \caption{Results on \benchmarkname{}: Small models ($\leq$ 3B parameters) show citation failure even in simple cases, while all models fail in more complex cases (see §\ref{sec:citation_analysis/results}). \reck{} / \reckf{}: Recall @$k$ on all instances / instances answered correctly. $\frac{|\mathcal{R}^{\mathrm{f}}|}{|\mathcal{R}|}$: Proportion of correctly answered instances.}  
    \label{tab:generation_all_results}
\end{table*}

We ran 18 LLMs\footnote{We omit ``-Instruct'' specifiers for brevity. See Table~\ref{tab:models} for model details} (5 families, 0.6B -- 120B) on \benchmarkname{}, showing results in Table~\ref{tab:generation_all_results}. After analyzing the effect of our proposed filtered evaluation, we interpret the results with regard to our main research question.

\paragraph{Citation performance is higher on correct responses} We observe that for most combinations of model and task, filtered recall @$k$ (\reckf{}) is higher than or equal to unfiltered evaluation (\reck{}). While the magnitude of this difference depends on the model and dataset, on \musique{} and \neoqa{} we observe up to $\sim$14 and $\sim$8 points difference between \reckf{} and \reck{}, respectively,\footnote{e.g. Qwen3-32B on \musique{}, Qwen3-1.7B on \neoqa{}} showing that differentiating between response failure and citation failure is important. In the following, we therefore focus on \reckf{} scores.

\paragraph{Small models fail for \single{} reasoning; all models fail for complex reasoning} On the \single{} reasoning datasets \repliqa{} and \boolqm{}, models with $\geq$3B parameters achieve near-perfect \reckf{} ($>95$), while smaller models score lower. On \musique{} and \neoqa{}, requiring \concatenation{} and \intersection{} reasoning over multiple documents, all models show reduced \reckf{} scores, indicating failure to identify evidence documents even though models correctly identify the necessary information for generating the response. Across models and tasks, there is a strong positive correlation between response scores\footnote{Answer F1 / Exact Match, data not shown.} and \reckf{} ($r=$0.745, $p=$0.000), showing that the response-evidence relation affects both response generation and citation.

\begin{figure*}
    \centering
    \includegraphics[]{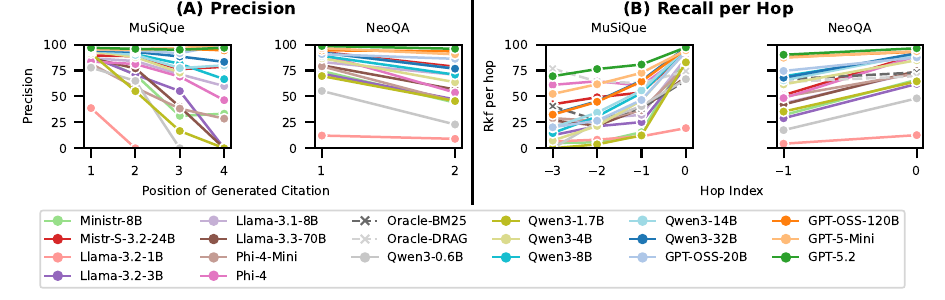}
    \includegraphics[]{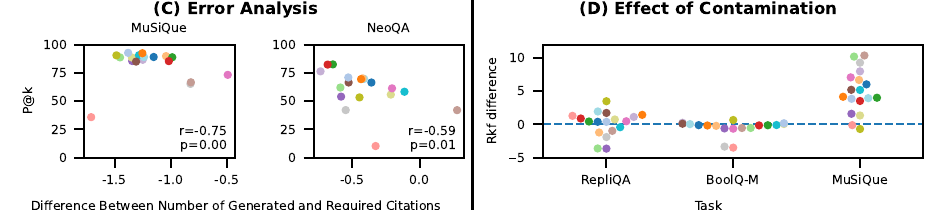}
    \caption{(A) Citation precision decreases with the order of appearance in generation, except for GPT models. (B) \reckf{} is highest for the evidence document that contains the response (hop 0) and is reduced when going to earlier hops, \drag{} and \bmr{} are notable exceptions. (C) Models tend to under-generate citations on \musique{} and \neoqa{}. Approaching the correct number of citations is correlated with lower precision ($r$ and $p$ exclude Llama-3.2-1B). (D) Contamination has diverse effects on citation performance. Plot shows difference in \reckf{} on contaminated / uncontaminated instances. See §\ref{sec:citation_analysis/results}.}
    \label{fig:generation_precision_recall_per_hop}
\end{figure*}

\paragraph{LLM citations are (imperfectly) ordered by confidence} Fig.~\ref{fig:generation_precision_recall_per_hop}(A) shows the precision of citations by their order of appearance in generation. It is visible that precision decreases from the first to the last citation for most models, suggesting that LLMs rank citations by confidence. The GPT models are notable exceptions, showing high precision on all generated citations.

\paragraph{Evidence recall decreases with moving up in the reasoning chain} Figure \ref{fig:generation_precision_recall_per_hop}(B) shows \reckf{} by evidence position in the reasoning chain for \musique{} and \neoqa{} (\concatenation{} instances only). Compared to hop 0 (\explicit{} overtness), \reckf{} drops strongly for earlier hops (\implicit{} overtness), showing that models struggle to trace the reasoning chain during citation. The retrieval-based baselines \drag{} and \bmr{} are notable exceptions, with higher recall at hop \mbox{$-$3} than hop \mbox{$-$2}. As retrieval-based models rely on both question and response to find evidence, their focus on the response is reduced. This explains their sub-optimal performance at hop 0 (which contains the response verbatim) but elevated scores at earlier hops, where the question provides helpful signals (visible in elevated ROUGE scores between question and evidence in Table~\ref{tab:rouge_scores}).


\paragraph{Implicitness alone does not entail citation failure} The response-evidence relation is \implicit{} in \boolqm{}, and \explicit{} in \repliqa{}, which explains the reduced \reckf{} for \bmr{} on \boolqm{}. In contrast, most LLMs exhibit \reckf{} scores close to 100 on \boolqm{}, showing that they are able to cite evidence in the absence of an \explicit{} statement-evidence relation. 

\paragraph{Explicitness can bias citation} In Fig.~\ref{fig:reckf_detailed_neoqa}, we show average \reckf{} on \concatenation{} and \intersection{} \neoqa{} instances, as well as \reckf{} per hop. As expected, for most LLMs, \reckf{} is highest on hop 0 evidence in \concatenation{} instances, likely due to the \explicit{} response-evidence relation. Unexpectedly, \reckf{} on hop \mbox{-1} and average \reckf{} on \concatenation{} instances is lower than average \reckf{} on \intersection{} instances for most models. This suggests that the \explicit{} relation between the response and hop 0 evidence in \concatenation{} instances biases models to cite only hop 0 correctly, while the absence of this bias allows for better average citation performance on \intersection{} instances. Again, \drag{} and \bmr{} are exceptions, as \reckf{} is higher on \concatenation{} instances than on \intersection{} instances.

\paragraph{Error analysis} The two main citation failure modes are (a) generating too few citations and (b) generating wrong citations. Table~\ref{tab:precision_vs_n_generated_citations} shows the number of instances with citation failure (\reckf{} $<$1), along with: cases without any citations ($n_\emptyset$), the average difference between generated and correct citation counts (Diff), and precision @$k$ (P@$k$). On the \single{} reasoning datasets \repliqa{} and \boolqm{}, failure is mostly due to wrong citations, with $n_\emptyset$ and Diff near 0. On \musique{} and \neoqa{} (Fig.~\ref{fig:generation_precision_recall_per_hop} C), both failure modes appear: most models generate some citations ($n_\emptyset=$0) but too few (Diff $<$ 0), and precision decreases as the number of citations grows. 

\paragraph{Data contamination has diverse effects on citation} Data contamination---the presence of relevant information in pretraining data \cite{sainz-etal-2023-nlp, li-etal-2024-open-source}---could reduce reliance on contextual information and thus hurt citation performance, though this has not been studied. Fig.~\ref{fig:generation_precision_recall_per_hop} (D) shows the difference in \reckf{} between contaminated and uncontaminated instances for \repliqa{}, \boolqm{}, and \musique{}.\footnote{For details, see §\ref{sec:replication/contamination}. We exclude \neoqa{} as contamination analysis is difficult given its multiple choice design.} On \repliqa{} and \boolqm{}, contamination has both beneficial and detrimental effects across models. On \musique{}, the effect is clearly beneficial, contradicting our initial expectation.

\paragraph{} Our analysis revealed that retrieval- and generation-based methods excel under different conditions. This suggests that combining citation methods can improve performance, which we will investigate in the following sections.

\section{\approachname{}: A Framework for Investigating Efficient LLM Citation}
\label{sec:approach}

All models showed citation failure in our experiments, calling for mitigation. Attention values are a promising direction for this purpose: They have been exploited successfully in related tasks \cite{chen2025attention, zhang2025queryfocusedretrievalheadsimprove} and can be used efficiently, requiring the training of only a small number of parameters (or none at all) and not requiring additional generation. However, their potential for citation has not been studied. 

To close this gap, we introduce \approachname{}, our framework for studying the attention-based enhancement of generative citation. After giving an overview, we describe the used citation methods in §\ref{sec:approach/methods} and their combination in §\ref{sec:approach/aggregation}.

\paragraph{Overview} To enhance generative citation with other efficient citation methods, we assume individual citation methods $\mathrm{M}(r, s) \rightarrow \mathbb{R}$, that predict a \textit{citation score} reflecting the relevance of document $s$ as evidence for response $r$. Our results in §\ref{sec:citation_analysis/results} and research on model ensembles \cite{dietterich2000ensembles} suggest that combining scores can improve performance, so we experiment with method combination $\mathrm{M}^{\Omega}=\mathrm{Agg}(\mathrm{M}^1,...,\mathrm{M}^{|\mathrm{M}^{\Omega}|})$, where $\mathrm{Agg}(\cdot)$ is an aggregation function and $|\mathrm{M}^{\Omega}|$ is the number of individual citation functions in $\mathrm{M}^{\Omega}$. Finally, we require a decision function $\delta(\mathrm{M}(r,s)):\mathbb{R} \rightarrow \{0,1\}$ that maps the citation score to a decision \texttt{no-cite} or \texttt{cite}.

\subsection{Citation Methods}
\label{sec:approach/methods}

To enhance generation-based citation, we consider the attention-based methods \icr{} \cite{chen2025attention}, \at{} \cite{cohenwang2024contextcite} and \qr{} \cite{zhang2025queryfocusedretrievalheadsimprove}. Retrieval-based methods, which have been applied successfully in related work on post-hoc citation (e.g. \citealt{bohnet2022attributed, sancheti-etal-2024-post}), serve as efficient baselines. We introduce these methods below and refer the reader to the respective publications for details.

\subsubsection{Generation-Based Citation}
\label{sec:approach/generation}

We obtain the citation score $\mathrm{M}^{\mathrm{Gen}}$ as the length-normalized \cite{murray-chiang-2018-correcting} probability for generating the citation (see §\ref{sec:replication/method_details/generation}). For source documents without citations, the score is 0.

\subsubsection{Attention-Based Citation}
\label{sec:approach/attention}

We focus on three recent attention-based methods: \icr{}, \cite{chen2025attention}, \textsc{QRhead} (\qr{}, \citealt{zhang2025queryfocusedretrievalheadsimprove}) and \at{} \cite{cohenwang2025learningattributeattention}. We first describe their general approach and then introduce the individual methods, adapting the notation from \citet{cohenwang2025learningattributeattention}. 

\paragraph{General approach} To obtain citation scores, the attention-based methods work in two steps: 
\begin{enumerate}
    \item Compute a single score $\mathrm{M}_d(r,s)$ per attention head $h_d$, where $d \in \{1,...,|H|\}$ and $|H|$ is the number of attention heads (see §\ref{sec:replication/method_details/attention}).
    \item Compute a weighted average over per-head scores: $\mathrm{M}(r,s)=\sum_{d=1}^{|H|}\theta_d \mathrm{M}_d(r,s)$.
\end{enumerate} 
The difference between the attention-based methods is in the way the weight vector $\theta$ is obtained:

\textbf{{\icr}} \cite{chen2025attention} puts equal weights on all attention heads,\footnote{\citet{chen2025attention} propose a calibration method for \icr{} that is also used in \qr{}. Our preliminary experiments showed that it leads to decreased performance, so we are not using it.} so $\forall d: \theta_d^{\mathrm{ICR}} = \frac{1}{|H|}$

\textbf{\qr{}} \cite{zhang2025queryfocusedretrievalheadsimprove} selects a subset of ``query-focused retrieval heads'' $H^{\mathrm{QR}} \subset H$:
\[
    \theta_d^{\mathrm{QR}} = 
    \begin{cases}
        \frac{1}{|H^{\mathrm{QR}}|} & \text{if } h_{d} \in H^{\mathrm{QR}} \\
        0 & \text{otherwise}
    \end{cases}
\]
$H^{\mathrm{QR}}$ is obtained as the heads that give the highest scores to relevant documents on a training set, where $|H^\mathrm{QR}|$ is set based on model size.

\textbf{\at{}} \cite{cohenwang2025learningattributeattention} learns a soft weighting $\theta^{\mathrm{AT2}}$ such that the score for a source document $s$ reflects the effect of removing $s$ from the context: For a given training example, an LLM generates a continuation. Source documents are removed randomly from the context, the change in probability of the original generation is recorded, and $\theta^{\mathrm{AT2}}$ is optimized with a correlation loss.

\begin{figure*}
    \includegraphics[trim={0 0.9cm 0 0},clip]{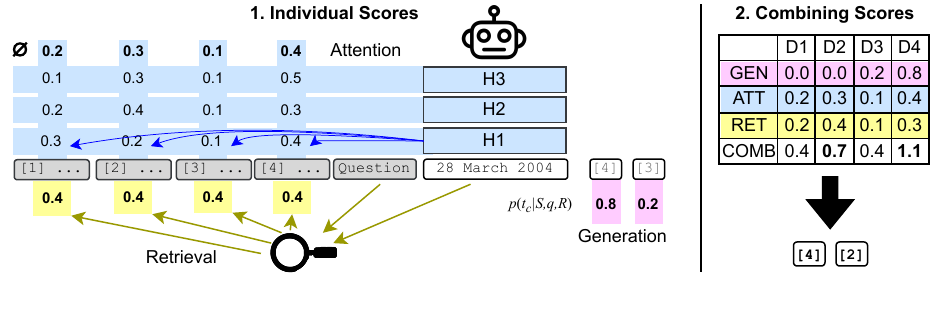}
    \caption{Overview of \approachname{} (§\ref{sec:approach}). Left: Individual scores for each document are obtained from generation-based, attention-based and retrieval-based methods. Attention scores are averaged over individual heads. Right: Scores from individual methods are summed to obtain a final citation prediction. Attention head weights $\theta$ and combination weights $w, b$ are omitted.}
    \label{fig:approach}
\end{figure*}

\subsubsection{Retrieval-Based Citation}
\label{sec:approach/retrievers}

As in §\ref{sec:citation_analysis}, we employ \bmr{} \cite{robertson2009bm25} and \drag{} \cite{lin-etal-2023-train}.

\subsection{Aggregation and Decision Functions}
\label{sec:approach/aggregation}

\paragraph{Aggregation} To aggregate scores from different citation methods, we use a weighted average:
\begin{equation}
 \mathrm{M}^{\Omega} = \sum_{i=1}^{|\mathrm{M}^\Omega|} w_i\mathrm{M}^i + b
\end{equation}

This retains efficiency and avoids introducing confounders into our analysis. To learn $w$ and $b$, we fit a linear model\footnote{Using a LinearModel from scikit-learn  \cite{scikit-learn}} on the train set scores from individual attention-based methods. We experiment with 3 combinations of scores: 
\begin{itemize}
    \item \comba{}: Generative and attention-based citation (\gen{}, \icr{}, \at{} and \qr{})
    \item \combr{}: Generative and retrieval-based citation (\gen{}, \bmr{} and \drag{})
    \item \comb{}: All \approachname{} methods (\gen{}, \icr{}, \at{}, \qr{}, \bmr{} and \drag{})
\end{itemize}

\paragraph{Decision Function} As done in previous work, we predict evidence by selecting the $k$ highest scoring source documents for a given response statement \cite{bohnet2022attributed, ramu-etal-2024-enhancing}. We choose $k$ depending on dataset and instance as described in §\ref{sec:replication/filtered_evaluation} and §\ref{sec:attention/transfer/setup}

\section{How Can Attention Values be Used to Efficiently Mitigate LLM Citation Failure?}
\label{sec:attention}

We study how to mitigate citation failures in generative citation using \approachname{}. Prior work has examined generation-based citation (e.g., \citealt{gao2023alce, tang2024lciteevallongcontextmodelstruly}) and retrieval-based citation (e.g., \citealt{bohnet2022attributed, ramu-etal-2024-enhancing}) separately, but we are the first to explore attention-based citation and to combine generation-, attention-, and retrieval-based approaches. To broaden the evaluation of \approachname{}, we complement experiments on \benchmarkname{} with a transfer setting using two challenging datasets from a recent long-document attribution benchmark \cite{buchmann-etal-2024-attribute}.

\subsection{Experimental Setup}
\label{sec:attention/setup}

\subsubsection{Methods} 

\paragraph{Training} Where available, we train on the train splits of the \benchmarkname{} datasets, and train on the dev split of \neoqa{}. We train one set of parameters per combination of LLM and task. For $\theta^\mathrm{QR}$, we randomly choose 150 examples per dataset for selecting heads, and set $|H^{\mathrm{QR}}|$ to 16 for models $\leq$8B parameters and else to 32 as in \citet{zhang2025queryfocusedretrievalheadsimprove}. For $\theta^\mathrm{AT2}$, we train on all available examples and use the same hyperparameters as in \citet{cohenwang2025learningattributeattention}. We train $w$ and $b$ for the combination methods on the train (dev) set scores of the individual \approachname{} methods. 

\paragraph{Masking Reasoning Tokens for Attention-Based Methods} Except where otherwise noted, we evaluate attention-based methods with masked reasoning tokens, as this improved their performance in preliminary experiments (see \ref{sec:attention/results/isolated}). 

\subsubsection{Transfer}
\label{sec:attention/transfer/setup}

\paragraph{Datasets} \qasper{} is a question-answering dataset on scientific articles proposed by  \citet{dasigi-etal-2021-dataset}. We exclude unanswerable instances. \govreport{} is a summarization dataset on reports from US government agencies proposed by \citet{huang-etal-2021-efficient}. Besides requiring longer response statements than the datasets in \benchmarkname{}, these datasets represent a shift in source document type (incoherent collection of paragraphs $\rightarrow$ coherent document) and for \qasper{} in domain (Wikipedia / General $\rightarrow$ scientific).

\paragraph{Evaluation} \qasper{} and \govreport{} are datasets with free-form responses and incomplete evidence annotations, requiring more flexible evaluation than \reck{}/\reckf{}. Therefore, we employ the attributability evaluation models Minicheck \cite{tang-etal-2024-minicheck} for \qasper{} and TRUE \cite{honovich2022true} for \govreport{}, which have been shown to obtain $>$75\% accuracy on these datasets \cite{buchmann-etal-2024-attribute}. For \qasper{}, we treat LLM responses as a single statement, while we split responses for \govreport{} by sentence using spacy.\footnote{\url{https://github.com/explosion/spaCy}} We report the proportion of response statements evaluated as attributable to the 2 highest-scoring source documents ($k=2$).

\paragraph{\approachname{} Parameters} Based on preliminary experiments, we use \at{} and \qr{} parameters trained on \musique{} for \qasper{}, and parameters trained on \repliqa{} for \govreport{}. We retrain $w$ and $b$ score combination parameters based on attributability scores. 

\subsection{Results}
\label{sec:attention/results}

\begin{figure*}[!t]
    \centering
    
    \includegraphics[]{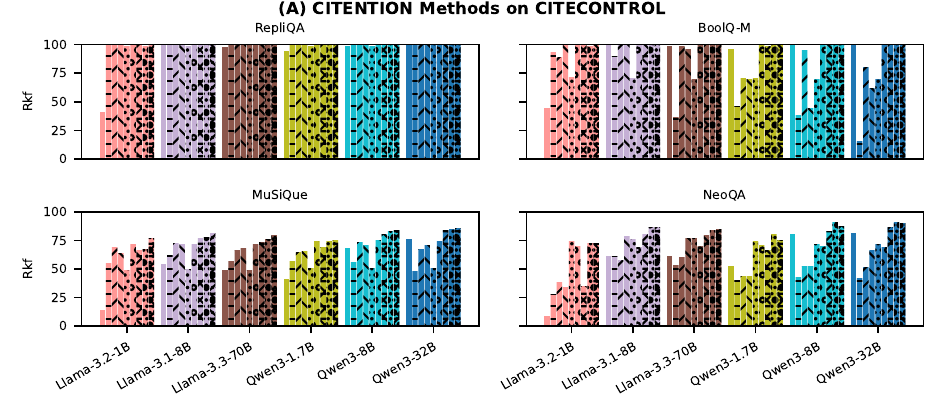}
    \includegraphics[trim={0 0.8cm 0 0},clip]{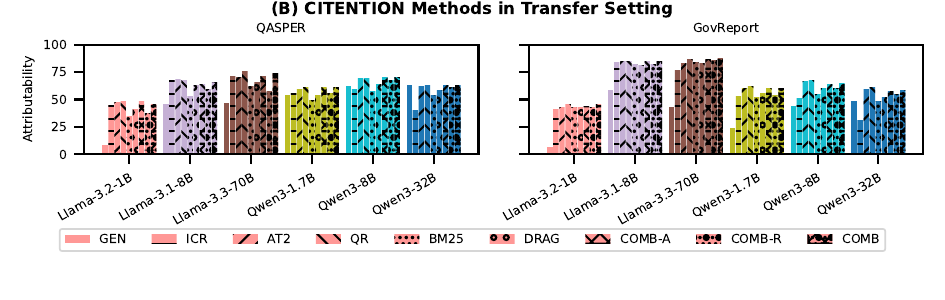}
    \caption{(A) \approachname{} methods improve citation \reckf{} scores on \benchmarkname{}. (B) \approachname{} methods trained on \benchmarkname{} improve \reckf{} scores on transfer tasks. Bars show proportion of answer statements that are attributable to the evidence. For numerical data see Table~\ref{tab:citention_all_results}.}
    \label{fig:attention_results}
\end{figure*}

We evaluated the performance of the \approachname{} methods on \benchmarkname{} and the transfer datasets, using Llama-3.2-1B, Llama-3.1-8B, Llama-3.3-70B, Qwen3-1.7B, Qwen3-8B and Qwen3-32B, showing results in Fig.~\ref{fig:attention_results} and Tab.~\ref{tab:citention_all_results}. We first analyze the performance of attention-based citation in comparison to generation-based citation and the retrieval baselines, and then the performance of method combination. 

\subsubsection{Results of Isolated Citation Methods}
\label{sec:attention/results/isolated}

\begin{figure}
    \centering
    \includegraphics[]{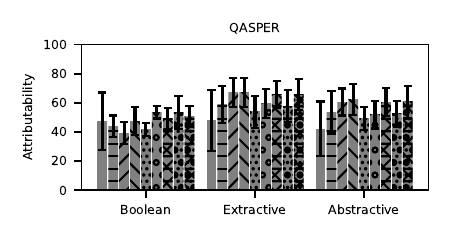}
    \caption{\reckf{} by question type on QASPER, averaged over models. Whiskers show standard deviation. See Fig.~\ref{fig:attention_results} for hatch legend.}
    \label{fig:qasper_answer_types}
\end{figure}

\paragraph{Attention-based and retrieval-based citation can mitigate citation failure} Attention-based and retrieval-based methods achieve higher average scores than generation-based citation (\gen{}) for the Llama models and Qwen3-1.7B on \benchmarkname{} (\drag{} improves over \gen{} by at least 6 points), and for all models on the transfer datasets (\qr{} improves over \gen{} by at least 5 points). This agrees with \citet{ramu-etal-2024-enhancing} and \citet{sancheti-etal-2024-post}, who found that retrieval-based methods can improve over LLMs in post-hoc citation, but did not investigate attention-based methods. In contrast, for Qwen3-8B and Qwen3-32B, \gen{} is stronger, as the average \reckf{} scores are more than 10 points above the average scores from \gen{} on the other models, and consistently higher than the scores from the other isolated citation methods. 

\paragraph{Retrieval improves over citation on \benchmarkname{}, attention-based methods are better on transfer datasets} On \benchmarkname{}, the average \reckf{} for \drag{} is mostly higher than for attention-based methods.  We attribute this to the fact that only the retrieval-based citation methods have access to the question, which is helpful in finding evidence with \implicit{} statement-evidence relation. This is visible in the ROUGE scores in Table~\ref{tab:rouge_scores}: There is consistent lexical overlap between the question and evidence for all datasets, while this is not the case for the response and the evidence. On the other hand, \qr{} and \at{} exhibit the highest scores on the transfer datasets. This suggests that the more long-form responses required for \qasper{} and \govreport{} enable the attention methods to use the LLM-internal information more effectively, giving them an advantage over retrieval-based methods that do not have access to this information.

\paragraph{Attention-based methods work better for Llama models than for Qwen models} We observe higher \reckf{} and attributability scores for attention-based methods on Llama models than on Qwen models. This indicates that attention-based citation is sensitive to model architecture and / or pretraining regime. 

\paragraph{Attention-based methods are sensitive to overtness} The sensitivity of attention-based methods to overtness can be seen in multiple places. First, while \reckf{} scores on \repliqa{} (\explicit{} statement-evidence relation) are consistently high for attention-based  methods, they are reduced on \boolqm{} (\implicit{}, Fig.~\ref{fig:attention_results}A). Similarly, attributability for extractive and abstractive questions is higher than for boolean questions on \qasper{} (Fig.~\ref{fig:qasper_answer_types}). Finally the performance of attention-based methods decreases when going from the evidence document that contains the response to the ones for earlier hops on \musique{} and \neoqa{} (Fig.~\ref{fig:attention_recall_per_hop_neoqa}). 

\begin{figure}
    \centering
    \includegraphics[trim={0 0.3cm 0 0},clip]{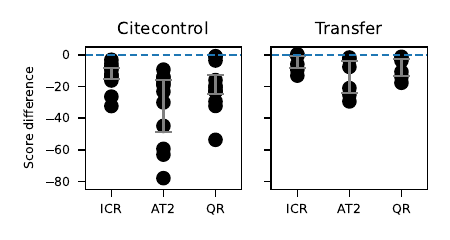}
    \caption{Attention-based citation suffers from reasoning tokens in context. Plots show difference between \reckf{} (\benchmarkname{}) / attributability (transfer) for attention based methods on Qwen models evaluated with and without reasoning tokens in context. Whiskers show interquartile range}
    \label{fig:reasoning_attention}
\end{figure}

\paragraph{\at{} and \qr{} provide better citations than \icr{}} In almost all combinations of model and task, \qr{} and \at{} achieve higher scores than \icr{}. We explain this with the fact that they were optimized on the respective datasets in \benchmarkname{}, while \icr{} was not.

\paragraph{Attention-based citation is distracted by reasoning tokens} Fig.~\ref{fig:reasoning_attention} shows the difference in \reckf{} / attributability between Qwen models evaluated with reasoning tokens in context and the same model evaluated with masked reasoning tokens. The negative values show that reasoning tokens hurt performance. All other plots and tables therefore only show scores from evaluation with masked reasoning tokens.

\subsubsection{Results of Combined Methods}
\label{sec:attention/results/combination}

\paragraph{Combining citation methods reliably mitigates citation failure.} The average citation score on \benchmarkname{} and transfer datasets is highest for all models when combining generation-based, attention-based and retrieval-based citation (\comb{}). For all combinations of models and datasets except Qwen3-32B on GovReport, \reckf{} / attributability is higher for \comb{} than for generation-based citation. For all models on \benchmarkname{}, average \reckf{} is higher when combining retrieval-based methods (\combr{}) than when combining attention-based methods (\comba{}), while it reversed on the transfer datasets. This reflects the performance of the isolated citation methods discussed above. 

\paragraph{Attention-based and retrieval-based citation can complement each other} The fact that the combination of citation methods improves over isolated methods suggests that they complement each other. This is especially visible in the recall of evidence by its position in the reasoning chain for Qwen3-8B and Qwen3-32B on \musique{} (Fig.~\ref{fig:attention_recall_per_hop_neoqa}). On hop 0, where the response-evidence relation is \explicit{}, we can observe the strongest improvements from attention-based citation. Going towards earlier hops (hops $-$1, $-$2, $-$3) with \implicit{} statement-evidence relation, we can observe that \drag{} exhibits the strongest performance, exemplifying how the different citation methods complement each other. 

\paragraph{Score combination requires complementary methods and is responsible to the response-evidence relation} While the combination of all citation methods (\comb{}) exhibits the highest average scores overall, it does not improve over individual citation methods on transfer datasets. To get more insights into the potential benefits of score combination, we propose to compute \textit{Oracle Improvement Bound} (OIB), which reflects the proportion of instances where the predicted citations from a weaker citation method can complement those from a stronger citation method (§\ref{sec:replication/details}). As visible in Fig.~\ref{fig:oracle_improvement_bound}, on \govreport{}, the highest observable OIB is considerably lower than on the other datasets except for \repliqa{}, where all methods obtain almost perfect scores. This indicates that there is not much complementary potential between citation methods on \govreport{}, and thus it is difficult for the combined methods to improve over isolated methods. \\
While OIB values on \qasper{} are higher than on \govreport{}, score combination does not always improve over isolated methods. As noted in §\ref{sec:attention/results/isolated}, the performance of the isolated citation methods varies between different response-evidence relation types on \qasper{}. As the score combination weights are constant, they cannot make ideal use of the citation methods depending on the relation type. This opens an exciting direction for research on improving the combination of citation methods.

\section{Conclusion}
\label{sec:conclusion}

We defined citation failure and presented key findings for understanding and mitigating it. Using \benchmarkname{}, we showed that citation failure is frequent, especially with complex statement-evidence relations. With \approachname{}, our results show that attention-based citation can mitigate failure from generative citation in some settings, suggesting that LLMs encode more than they generate, and making attention-based citation a promising research direction. More broadly, our analysis of attention-based citation can inform research on attention-based reranking.

Further, we demonstrated that citation failure can be reliably mitigated by combining generative, retrieval-based, and attention-based citation methods. Our work paves the way for practical, efficient citation methods, particularly in resource-constrained settings, and motivates future research on more sophisticated approaches to combine citation methods.

\section*{Acknowledgements}

This work has been funded by the LOEWE Distinguished Chair “Ubiquitous Knowledge Processing”, LOEWE initiative, Hesse, Germany (Grant Number: LOEWE/4a//519/05/00.002(0002)/81) and the European Union (ERC, InterText, 101054961). Views and opinions expressed are however those of the authors only and do not necessarily reflect those of the European Union or the European Research Council. Neither the European Union nor the granting authority can be held responsible for them. We express our sincere gratitude to the reviewers and action editor at TACL for their valuable and
constructive feedback. 

\bibliography{main}
\bibliographystyle{acl_natbib}

\onecolumn
\appendix

\section{Replication Information}
\label{sec:replication}

\subsection{Prompts}
\label{sec:replication/prompts}

We use a 3-shot prompt per dataset with diverse in-context examples and shortened source documents. Each source document $s_i \in S$ is prepended by its index in square brackets \cite{gao2023alce}, with questions placed after the source documents \cite{buchmann-etal-2024-attribute}. Task and format explanations are in Tables~\ref{tab:task_explanations} and \ref{tab:format_explanations}.

\begin{tcolorbox}[
  colframe=black!50,   
  boxrule=0.5pt,      
  arc=6pt,            
  left=6pt,           
  right=6pt,          
  width=\linewidth,  
]
{\ttfamily
<user\_input\_start> \\
\textcolor{blue}{
\# Task Explanation \\
Task: \{task\_explanation\} \\
}

\textcolor{teal}{
\# Format Explanation \\
Follow this example for answer formatting: \\
\{format\_explanation\} \\
}

\# 3-shot examples omitted\\

<user\_input\_start> \\
Retrieved Paragraphs: [0] \{document\_0\} \\
{[1]} \{document\_1\} \\
... \\

Question: \{question\} \\

<assistant\_input\_start> \\
Answer:\_
}
\end{tcolorbox}

\begin{table}
    \small
    \begin{tabular}{l|p{12cm}}
\toprule
\textbf{Dataset} & \textcolor{blue}{\textbf{task\_explanation}} \\
\midrule
\repliqa{} & You are given a question and a list of retrieved paragraphs, which might contain relevant information to the question. Answer the Question using only the information from the retrieved paragraphs. Your answer should be concise and not more than a single phrase. Do not provide any explanation. Provide the paragraph that can be used to verify the answer by writing the integer id in square brackets. \\
\midrule
\boolqm{} & You are given a yes/no question and a list of retrieved paragraphs, which might contain relevant information to the question. Answer the Question using only the information from the retrieved paragraphs. Your answer should be "yes" or "no". Do not provide any explanation. Provide the paragraph that can be used to verify the answer by writing the integer id in square brackets. \\
\midrule
\musique{} & You are given a question and a list of retrieved paragraphs, which might contain relevant information to the question. Answer the Question using only the information from the retrieved paragraphs. Your answer should be concise and not more than a single phrase. Do not provide any explanation. Provide all paragraphs that are needed to verify the answer by writing the integer id in square brackets. \\
\midrule
\neoqa{} & You are given a list of retrieved news articles, a question and 6 answer options. The news articles might contain relevant information to the question. Answer the Question by responding with one of the answer options. Your answer should be exactly the same as one of the answer options. Do not provide any explanation. Provide the ids of the news articles that information needed to answer the question by writing the integer id in square brackets. \\
\midrule
\qasper{} & You are given a Scientific Article and a Question. Answer the Question as concisely as you can, using a single phrase or sentence. If the question is a yes/no question, your answer should be "yes" or "no". Do not provide any explanation. Provide the paragraphs that can be used to verify the answer by writing their integer ids in square brackets. Put each id in separate brackets. Always provide ids of content paragraphs, not section headlines. \\
\midrule
\govreport{} & Task: You are given a government report document. Write a one page summary (max. 15 sentences) of the document. Each sentence in your summary should reference the source paragraphs from the document that can be used to verify the summary sentence. There should be no sentences without references. Always put the references after the summary sentence. \\
\bottomrule
\end{tabular}
    \caption{Task explanations used in the prompts in our experiments.}
    \label{tab:task_explanations}
\end{table}

\begin{table}
    \small
    \begin{tabular}{l|p{12cm}}
\toprule
\textbf{Dataset} & \textcolor{teal}{\textbf{format\_explanation}} \\
\midrule
\repliqa{} & Retrieved Paragraphs: <omitted> \newline
Question: When did Beyonce start becoming popular? \newline
Answer: in the late 1990s [7]
 \\
\midrule
\boolq{} / \newline \boolqm{} & Retrieved Paragraphs: <omitted> \newline
Question: is axl rose the new lead singer of acdc \newline
Answer: yes \\
\midrule
\musique{} & Retrieved Paragraphs: <omitted> \newline
Question: When was the institute that owned The Collegian founded? \newline
Answer: 1960 [5] [9]  \\
\midrule
\neoqa{} & News Articles: <omitted> \newline
Question: What is the duration between the date when Crestfield Property Holdings shared the preliminary findings of the ZentroTek Solutions review (assumed to be shared by the end of January) and the date when Everstead Technical Systems discovered the calibration issue affecting the surveillance cameras? \newline
Answer options: a) 21 days \newline
b) 35 days \newline
c) 30 days \newline
d) 31 days \newline
e) 28 days \newline
f) 14 days \newline
Answer: 28 days [4] [7] \\
\midrule
\qasper{} & Scientific Article: <omitted> \newline
Question: Which baselines were used? \newline
Answer: BERT, RoBERTa [7] [8] \\
\midrule
\govreport{} & Report: <omitted> \newline
Under the Arms Export Control Act and its implementing regulations, DOD is required to recover nonrecurring costs—unique one-time program-wide expenditures—for certain major defense equipment sold under the FMS program. [2] [4] These costs include research, development, and one-time production costs, such as expenses for testing equipment. [3] \\
\bottomrule
\end{tabular}
    \caption{Format explanations used in the prompts in our experiments.}
    \label{tab:format_explanations}
\end{table}

\begin{table*}[]
\small
    \centering
    \begin{tabular}{l|l}
    \textbf{Purpose}     &  \textbf{Package} \\
    \midrule
    Base for \approachname{}     & AT2 \cite{cohenwang2025learningattributeattention} \\
    Generation & Huggingface Transformers \cite{wolf2020huggingfacestransformersstateoftheartnatural} \\
    BM25 retrieval & Rank-BM25\footnote{\url{https://pypi.org/project/rank-bm25/}} \\
    Dense retrieval & Sentence Transformers \cite{reimers-2019-sentence-bert} \\
    Aggregation weight fitting & Scikit-Learn \cite{scikit-learn} \\
    ROUGE score computation & Rouge-Score\footnote{\url{https://pypi.org/project/rouge-score/}}
    \end{tabular}
    \caption{Python packages used in experiments.}
    \label{tab:packages}
\end{table*} 

\subsection{Data Processing}
\label{sec:replication/data_processing}

\textbf{\repliqa{}} \cite{montciro-nips-2024-repliqa} is a dataset of human-written multi-paragraph documents, questions, long answers and short answers. We only use questions where both answers appear verbatim in the context document. We split the context documents split into paragraphs and sampled one answer-containing paragraph as evidence. We sampled 19 distractor paragraphs from the same document (non-answer paragraphs) and supplemented them with paragraphs from other documents if needed.

\textbf{\boolqm{}} was created in a two-step modificiation process of \boolq{} \cite{clark-etal-2019-boolq}, which is a dataset of context passages, questions and yes/no answers: (1) \textbf{Filtering}: GPT-OSS-120B selected instances where the question concerns a concrete, identifiable entity that appears in the passage (excluding questions about abstract concepts, laws, or definitions). (2) \textbf{Entity replacement}: GPT-OSS-120B  replaced entities in suitable instances with fictional ones, being instructed to preserve the question and passage structure and style, logic, and original truth value, and to not use replacements from existing fictional universes. Distractor source documents for \boolq{} / \boolqm{} are 19 randomly sampled paragraphs from other instances in the same split.

\textbf{\musique{}} \cite{trivedi-etal-2022-musique} is a dataset of questions, answers, 2 -- 4 relevant context paragraphs and 16 -- 18 distractor paragraphs, questions and answers, which we used as is. 

\textbf{\neoqa{}} \cite{glockner2025neoqaevidencebasedquestionanswering} is a dataset of fictional news articles about one of several event timelines, questions, and 6 answer options (one of them true). To assemble source documents for a specific question, we sampled 2 news articles that contain the relevant information pieces as evidence, and sampled 18 news articles from the same timeline that do not contain the relevant information pieces as distractors. 

\paragraph{Dataset splits} \repliqa{} was sequentially released in 5 splits between June 2024 and June 2025\footnote{\url{https://github.com/ServiceNow/repliqa}}. We use splits 0--2 for training, split 3 for development and the latest split 4 for testing. We use the development splits of, \boolq{} / \boolqm{} and \musique{} for evaluation, as their test splits are hidden. We use the test splits of \neoqa{}, \qasper{} and \govreport{} (CRS subset). 

\subsection{Filtered evaluation (\reckf{})} 
\label{sec:replication/filtered_evaluation}

To perform filtered evaluation, we consider responses with a response evaluation score $>$0.7. To evaluate response correctness, we use token F1 score for \repliqa{} and \musique{}, and exact match for \boolq{}, \boolqm{} and \neoqa{}, as done in the respective original dataset papers. We set $k = |E^*| + 1$, i.e. one larger than the size of the ground truth evidence set for a particular instance. We assume the order of generated citations as their ranking from highest to lowest.

\subsection{Comparison between contaminated and uncontaminated instances}
\label{sec:replication/contamination}

For \repliqa{} and \musique{}, we determine the contamination status of test set questions for a specific model by prompting without source documents. If the answer F1 score is (a) above 0.7 (b) below 0.3 (c) in between, the question is (a) seen as contaminated (b) uncontaminated (c) discarded for this analysis. \\
As the chance for randomly answering a \boolqm{} question correctly is 50\%, determining the contamination status of individual questions is difficult. Instead, we assume all original \boolq{} questions as contaminated, and our newly created \boolqm{} as uncontaminated. This is verified in Table~\ref{tab:memorization}, which shows that performance on \boolqm{} without context is strongly reduced and below random chance, while it is above random chance and higher for \boolq{} without context.

\subsection{Employed LLMs}
\label{sec:replication/llms}

\begin{table*}
    \tiny
    \centering
    \begin{tabular}{lllll}
    \toprule
    Model & \#Params \newline (Active) & Reasoning Type & Citation & Huggingface Tag \\
    \midrule
    Ministr-8B & 8B & No Reasoning & \citealt{liu2026ministral3} & \texttt{mistralai/Ministral-8B-Instruct-2410} \\
    Mistr-S-3.2-24B & 24B & No Reasoning & \href{https://huggingface.co/mistralai/Mistral-Small-3.2-24B-Instruct-2506}{Huggingface} & \texttt{mistralai/Mistral-Small-3.2-24B-Instruct-2506} \\
    Llama-3.2-1B & 1B & No Reasoning & \citealt{grattafiori2024llama3herdmodels} & \texttt{meta-llama/Llama-3.2-1B-Instruct} \\
    Llama-3.2-3B & 3B & No Reasoning & \citealt{grattafiori2024llama3herdmodels} & \texttt{meta-llama/Llama-3.2-3B-Instruct} \\
    Llama-3.1-8B & 8B & No Reasoning & \citealt{grattafiori2024llama3herdmodels} & \texttt{meta-llama/Llama-3.1-8B-Instruct} \\
    Llama-3.3-70B & 70B & No Reasoning & \citealt{grattafiori2024llama3herdmodels} & \texttt{meta-llama/Llama-3.3-70B-Instruct} \\
    Phi-4-Mini & 4B & No Reasoning & \citealt{abdin2024phi4technicalreport} & \texttt{microsoft/phi-4-mini-instruct} \\
    Phi-4 & 14B & No Reasoning & \citealt{abdin2024phi4technicalreport} & \texttt{microsoft/phi-4} \\
    Qwen3-0.6B & 0.6B & Reasoning & \citealt{yang2025qwen3technicalreport} & \texttt{Qwen/Qwen3-0.6B} \\
    Qwen3-1.7B & 1.7B & Reasoning & \citealt{yang2025qwen3technicalreport} & \texttt{Qwen/Qwen3-1.7B} \\
    Qwen3-4B & 4B & Reasoning & \citealt{yang2025qwen3technicalreport} & \texttt{Qwen/Qwen3-4B} \\
    Qwen3-8B & 8B & Reasoning & \citealt{yang2025qwen3technicalreport} & \texttt{Qwen/Qwen3-8B} \\
    Qwen3-14B & 14B & Reasoning & \citealt{yang2025qwen3technicalreport} & \texttt{Qwen/Qwen3-14B} \\
    Qwen3-32B & 32B & Reasoning & \citealt{yang2025qwen3technicalreport} & \texttt{Qwen/Qwen3-32B} \\
    GPT-OSS-20B & 20B (3.6B) & Reasoning & \citealt{openai2025gptoss120bgptoss20bmodel} & \texttt{openai/gpt-oss-20b} \\
    GPT-OSS-120B & 120B (3.6B) & Reasoning & \citealt{openai2025gptoss120bgptoss20bmodel} & \texttt{openai/gpt-oss-120b} \\
    GPT-5-mini & N/A & Reasoning & \href{https://developers.openai.com/api/docs/models/gpt-5-mini}{OpenAI Website} & N/A \\
    GPT-5.2 & N/A & Reasoning & \href{https://developers.openai.com/api/docs/models/gpt-5.2}{OpenAI Website} & N/A \\
    \bottomrule
\end{tabular}
    \caption{The LLMs employed in this work.}
    \label{tab:models}
\end{table*}

\subsection{Details of Citation Methods in \approachname{}}
\label{sec:replication/method_details}

\subsubsection{Generation-Based Citation}
\label{sec:replication/method_details/generation}

For generation-based citation, we obtain the citation score for source document $s_j$ as the length-normalized \cite{murray-chiang-2018-correcting} probability for generating the citation tokens $c = \{t^{
c}_1...t^{c}_{|c|}\}$ that point to $s_j$ (e.g. "\texttt{[4]}"). 

\begin{equation}
\begin{split}
\mathrm{M}^{\mathrm{Gen}}(r, s) = \\
 \exp\left( \frac{1}{|c|} \sum_{i=1}^{|c|} \log \left( \frac{\exp(\ell_t[t^{c}_i])}{\sum_{v=1}^{V} \exp(\ell_t[v])} \right) \right)
\end{split}
\end{equation}

which is equivalent to the geometric mean of the token probabilities:

\begin{equation}    
\mathrm{M}^{\mathrm{Gen}}(r, s) = \left( \prod_{i=1}^{|c|} p(t^{c}_i \mid x,  t^{c}_{<i}) \right)^{1/|c|}
\end{equation}

\subsubsection{Attention-Based Citation}
\label{sec:replication/method_details/attention}

\paragraph{Computing per-head attention scores} The per-head attention score $\mathrm{M}_{d}(r, s)$, for query $r$ consisting of tokens $t^r_{1}...t^r_{|r|}$ and source document $s$ consisting of tokens $t^s_{1}...t^s_{|s|}$ is obtained as
\begin{equation}
    \mathrm{M}_{d}(r, s)  = \frac{1}{|r|}\sum_{i=1}^{|r|} \sum_{j=1}^{|s|} \mathrm{ATT}_d(t^r_i, t^s_j)
\end{equation}
$\mathrm{ATT}_d(t_i, t_j)$ is the softmax-normalized attention score from token $i$ to token $j$ in head $h_d$.

\subsubsection{Retrieval-Based Citation}
\label{sec:replication/method_details/retrieval}

\paragraph{\bmr{}} \bmr{} \cite{robertson2009bm25} computes relevance scores by computing the token overlap between query and document, and weighting overlapping tokens according to their frequency of occurence in a training corpus. While computationally simple, it is still considered a competitive retrieval baseline \cite{thakur2021beir}. We compute token frequency statistics on the train sets of the \benchmarkname{} tasks\footnote{We use the dev set of \neoqa{} as it does not have a train set.} and set hyperparameters to common values \texttt{k1}$=1.5$; \texttt{b}$=0.75$ \cite{robertson2009bm25}.

\paragraph{\drag{}} Dragon \cite{lin-etal-2023-train} is based on a dual transformer-encoder architecture and was trained with a mixture of data augmentation techniques. Relevance scores are computed as the dot product of the query and document vector representations.  We leave the parameters unchanged, as it has been optimized for zero-shot retrieval.

\subsection{Technical Details}
\label{sec:replication/details} 

\begin{figure*}[!htb]
    \centering
    \includegraphics[trim={0 0.5cm 0 0},clip]{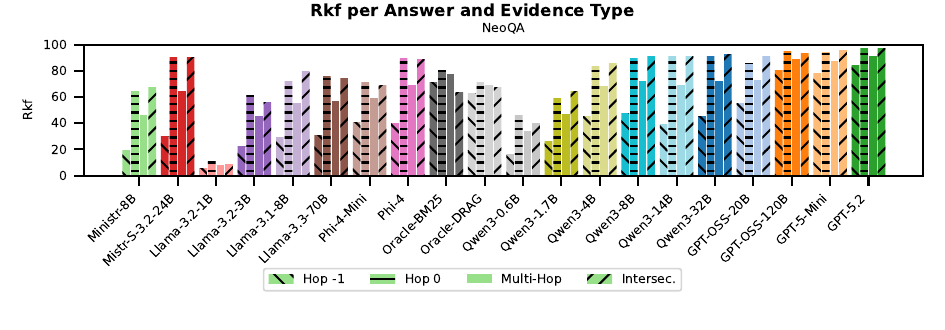}
    \caption{Detailed results for \neoqa{}. Hop -1/0: \reckf{} per hop on \concatenation{} instances. \concatenation{} / \intersection{}: average \reckf{} for the respective instance type. For analysis and discussion see §\ref{sec:citation_analysis}.}
    \label{fig:reckf_detailed_neoqa}
\end{figure*}

\paragraph{Mapping evidence to hops in NeoQA} To perform the analysis of recall per hop for NeoQA (Figs. \ref{fig:generation_precision_recall_per_hop} and \ref{fig:reckf_detailed_neoqa}), a mapping between the evidence documents and the hop index is needed. To obtain this mapping, we ordered the evidence documents by lexical overlap (ROUGE-1) with the ground truth answer. The document with the higher ROUGE-1 was used as the evidence for hop 0, while the document with the lower ROUGE-1 was used as the evidence for hop -1.

\paragraph{Oracle Improvement Bound}
On the recall-focused \benchmarkname{} tasks, the highest possible recall from a combination of two citation methods $\mathrm{M}^a$ and $\mathrm{M}^b$ with predicted citation sets $E^a_i$ and $E^b_i$ on instance $i$ can be obtained by replacing incorrect predictions of one method with correct predictions from the other, assuming oracle knowledge of the ground truth and a fixed maximum number of citations. Let $E^{b \rightarrow a}_i$ denote the set of additional correct citations that method $a$ could obtain from $b$, and let $E^*_i$ be the set of ground-truth citations.

The recall-based directional potential improvement ratio for improving $a$ using $b$ is defined as
\[
I^R_{b \rightarrow a} = 
\frac{\sum_i |E^{b \rightarrow a}_i|}
     {\sum_i |E^*_i|}.
\]

Analogously, $I_{a \rightarrow b}$ is defined by exchanging the roles of $a$ and $b$.

For the transfer tasks, we do not have knowledge of ground truth citations. Thus, we set $$E^{b \rightarrow a}_i$$ to 1 if and only if the response statement $ri$ is attributable to $E^b_i$, but not to $E^a_i$, and vice versa. Then we compute 
\[
I^{\mathrm{Att}}_{b \rightarrow a} = 
\frac{\sum_i |E^{b \rightarrow a}_i|}
     {|\mathcal{R}|}.
\]
Since the two methods may differ in their baseline recall, these directional improvement ratios are generally asymmetric. When estimating the maximal achievable improvement of the better-performing method under a fixed prediction budget, only the smaller of the two ratios constitutes a valid upper bound. The larger ratio corresponds to improving the weaker method and therefore overestimates the achievable gain when starting from the stronger system. We therefore define the Oracle Improvement Bound (OIB) as
\[
\mathrm{OIB} = \min\left(I_{b \rightarrow a},\, I_{a \rightarrow b}\right).
\]

\paragraph{Used packages} See Tab.~\ref{tab:packages}

\subsection{Limitations}

\paragraph{Analysis of Source document composition} Several works analyze the effect of source document composition on citation performance (§\ref{sec:related_work}), which we view as complementary to our response-evidence analysis. Due to the large number of experiments, we fixed source documents to one set per instance. Future work could investigate interactions between both dimensions using \benchmarkname{} as a starting point.

\paragraph{Score combination method} We use weighted averages to combine citation scores, which is efficient and avoids confounders. While we showed this improves over isolated citation methods, it represents a lower bound for score combination. More sophisticated combination methods could further improve performance, and we hope future work builds on \approachname{} to explore this direction.

\paragraph{Closed source models} Attention-based citation requires access to model internals, precluding closed-source models. While our extensive experiments demonstrate that attention-based citation can efficiently enhance citation performance, future analysis on closed-source models would be valuable.

\clearpage
\newpage

\section{Additional Results}
\label{sec:additional_results}

\begin{table*}[!htb]
    \centering
    \small
    \begin{tabular}{ll|cc|cc}
\toprule
 &  & \multicolumn{2}{c}{Answer} & \multicolumn{2}{c}{Question} \\
 &  & Evi & No Evi & Evi & No Evi \\
\midrule
RepliQA & - & 1.00 & 0.18 & 0.75 & 0.31 \\
BoolQ-M & - & 0.03 & 0.02 & 0.72 & 0.27 \\
MuSiQue & Multi-Hop & 0.07/0.08/0.12/0.97 & 0.10 & 0.36/0.40/0.46/0.44 & 0.43 \\
\multirow{2}{*}{NeoQA} & Multi-Hop & 0.48/0.96 & 0.43 & 0.71/0.65 & 0.50 \\
 & Intersection & 0.17 & 0.14 & 0.67 & 0.54 \\
\multirow{3}{*}{QASPER} & Boolean & 0.01 & 0.00 & 0.41 & 0.18 \\
 & Extractive & 0.85 & 0.14 & 0.44 & 0.20 \\
 & Abstractive & 0.63 & 0.17 & 0.43 & 0.20 \\
\bottomrule
\end{tabular}

    \caption{Lexical overlap between ground truth answers / questions and evidence / no evidence source documents. Numbers show ROUGE-1 recall for tokens from answers / questions in evidence documents. For \concatenation{} instances, we show values per evidence document for hop .../-1/0.}
    \label{tab:rouge_scores}
\end{table*}

\begin{table}[]
    \centering
    \small
    \begin{tabular}{l|cc|cc|cc|cc|cc}
\toprule
 & \multicolumn{2}{c|}{\textbf{\repliqa{}}} & \multicolumn{2}{c|}{\textbf{\boolq{}}} & \multicolumn{2}{c|}{\textbf{\boolqm{}}} & \multicolumn{2}{c|}{\textbf{\musique{}}} & \multicolumn{2}{c}{\textbf{\neoqa{}}} \\ 
  & C & NC & C & NC & C & NC & C & NC & C & NC \\
\midrule
Llama-3.3-70B & 86.2 & 15.3 & 76.5 & 62.6 & 78.4 & 43.4 & 37.9 & 12.0 & 52.8 & 33.7 \\
Qwen3-32B & 87.0 & 17.7 & 90.5 & 75.2 & 90.0 & 41.0 & 61.8 & 17.9 & 84.8 & 10.4 \\
GPT-OSS-120B & 85.3 & 18.5 & 91.6 & 55.4 & 90.0 & 30.4 & 70.2 & 24.9 & 86.0 & 38.5 \\
GPT-5-Mini & 87.6 & 14.4 & 89.4 & 75.3 & 89.2 & 44.1 & 71.8 & 34.9 & 88.1 & 47.0 \\
GPT-5.2 & 88.2 & 16.0 & 91.3 & 83.3 & 90.6 & 45.2 & 75.8 & 36.0 & 96.1 & 46.7 \\
\bottomrule
\end{tabular}
    \caption{Data contamination on \benchmarkname{}: Response performance with (C) and without context (NC). Metrics: \repliqa{}, \musique{}: Answer F1. \boolq{}, \boolqm{}, \neoqa{}: Accuracy}
    \label{tab:memorization}
\end{table}

\begin{figure*}[!htb]
    \centering
    \includegraphics[trim={0 0.4cm 0 0.4cm},clip]{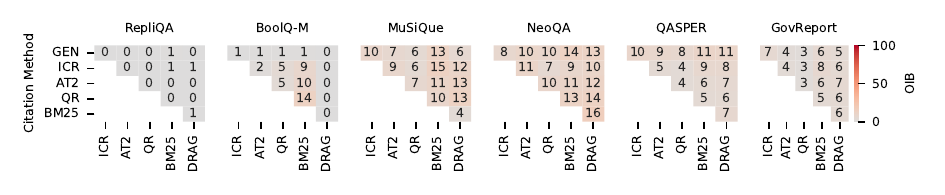}\label{fig:max_improvement_ratio}
    \caption{Oracle improvement bounds (OIB) averaged over models. See §\ref{sec:attention/results/combination} and §\ref{sec:replication/details} for details.}
    \label{fig:oracle_improvement_bound}
\end{figure*}

\begin{table}[]
    \centering
    \small
    \begin{tabular}{l|ccc|ccc|cccc|cccc}
\toprule
 & \multicolumn{3}{c|}{\repliqa{}} & \multicolumn{3}{c|}{\boolqm{}} & \multicolumn{4}{c|}{\musique{}} & \multicolumn{4}{c}{\neoqa{}} \\
 & $n$ & $n_\emptyset$ & Diff & $n$ & $n_\emptyset$ & Diff & $n$ & $n_\emptyset$ & Diff & P@$k$ & $n$ & $n_\emptyset$ & Diff & P@$k$ \\
\midrule
Ministr-8B & 26 & 0 & 0.0 & 3 & 0 & 0.0 & 597 & 1 & -1.5 & 88.7 & 421 & 2 & -0.6 & 62.0 \\
Mistr-S-3.2-24B & 1 & 0 & 0.0 & 1 & 1 & -1.0 & 749 & 0 & -1.0 & 85.4 & 427 & 0 & -0.7 & 82.3 \\
Llama-3.2-1B & 772 & 98 & -0.1 & 703 & 0 & 0.0 & 278 & 20 & -1.7 & 35.8 & 220 & 9 & -0.3 & 10.2 \\
Llama-3.2-3B & 44 & 9 & -0.2 & 21 & 0 & 0.0 & 555 & 0 & -1.3 & 85.5 & 370 & 13 & -0.6 & 54.0 \\
Llama-3.1-8B & 7 & 0 & 0.0 & 12 & 0 & 0.0 & 734 & 0 & -1.3 & 86.6 & 433 & 0 & -0.7 & 76.3 \\
Llama-3.3-70B & 30 & 0 & 0.0 & 20 & 0 & 0.0 & 633 & 2 & -1.3 & 85.0 & 413 & 0 & -0.5 & 66.5 \\
Phi-4-Mini & 66 & 3 & -0.0 & 33 & 0 & 0.0 & 429 & 1 & -0.8 & 66.6 & 316 & 0 & 0.3 & 42.0 \\
Phi-4 & 13 & 0 & 0.0 & 11 & 0 & 0.0 & 481 & 0 & -0.5 & 73.2 & 302 & 0 & -0.2 & 61.3 \\
Qwen3-0.6B & 49 & 0 & 0.0 & 153 & 1 & -0.0 & 23 & 0 & -0.8 & 65.2 & 373 & 12 & -0.5 & 42.0 \\
Qwen3-1.7B & 81 & 0 & 0.0 & 69 & 0 & 0.0 & 698 & 0 & -1.5 & 90.5 & 407 & 2 & -0.4 & 53.3 \\
Qwen3-4B & 15 & 0 & 0.0 & 7 & 0 & 0.0 & 562 & 0 & -1.3 & 88.6 & 350 & 0 & -0.2 & 55.8 \\
Qwen3-8B & 23 & 0 & 0.0 & 4 & 0 & 0.0 & 761 & 0 & -1.3 & 90.6 & 328 & 0 & -0.1 & 58.3 \\
Qwen3-14B & 6 & 0 & 0.0 & 6 & 0 & 0.0 & 744 & 0 & -1.2 & 89.0 & 368 & 0 & -0.4 & 69.7 \\
Qwen3-32B & 4 & 0 & 0.0 & 1 & 0 & 0.0 & 679 & 0 & -1.2 & 89.0 & 346 & 0 & -0.4 & 66.4 \\
GPT-OSS-20B & 39 & 0 & 0.0 & 7 & 0 & 0.0 & 805 & 1 & -1.4 & 92.7 & 215 & 0 & -0.5 & 70.9 \\
GPT-OSS-120B & 7 & 0 & 0.0 & 0 & 0 & 0.0 & 825 & 1 & -1.3 & 92.4 & 170 & 0 & -0.4 & 69.4 \\
GPT-5-Mini & 7 & 0 & 0.0 & 1 & 0 & 0.0 & 718 & 0 & -1.0 & 89.8 & 179 & 0 & -0.4 & 70.1 \\
GPT-5.2 & 0 & 0 & 0.0 & 0 & 0 & 0.0 & 504 & 0 & -1.0 & 88.8 & 138 & 0 & -0.6 & 82.5 \\
\bottomrule
\end{tabular}
    \caption{Citation error analysis: Table shows number of instances where \reckf{} is lower than 100 ($n$) as well as the average difference to the required number of citations (Diff) and precision@$k$ on these instances (P@$k$, omitted for \repliqa{} and \boolqm{} as it is always 0). See §\ref{sec:citation_analysis} for details}
    \label{tab:precision_vs_n_generated_citations}
\end{table}

\begin{table*}[!htb]
    \small
    \centering
    \begin{tabular}{ll|cccc|c||cc|c||c}
\toprule
 & & \multicolumn{5}{c||}{\textbf{\benchmarkname{}}} & \multicolumn{3}{c||}{\textbf{Transfer}} & \\
 & & \textbf{Re} & \textbf{Bo} & \textbf{Mu} & \textbf{Ne} & \textbf{Avg} & \textbf{QA} & \textbf{GO} & \textbf{Avg} & \textbf{Avg} \\
\midrule
\multirow[t]{9}{*}{Llama-3.2-1B} & GEN & 41.4 & 44.6 & 14.1 & 8.6 & 27.2 & 8.4 & 7.1 & 7.7 & 19.9 \\
 & ICR & 99.8 & 93.0 & 54.4 & 27.4 & 68.6 & 44.5 & 40.9 & 42.7 & 58.9 \\
 & AT2 & 99.8 & 89.0 & 68.9 & 38.0 & 73.9 & 46.9 & 42.9 & 44.9 & 63.0 \\
 & QR & \textbf{99.9} & 99.8 & 63.8 & 34.2 & 74.4 & \textbf{48.4} & \textbf{45.5} & \textbf{46.9} & 64.1 \\
 & BM25 & 98.1 & 71.7 & 48.3 & \textbf{73.8} & 72.9 & 34.5 & 42.6 & 38.5 & 60.0 \\
 & DRAG & 99.8 & \textbf{100.0} & 71.3 & 69.5 & 85.1 & 40.8 & 42.7 & 41.7 & 68.8 \\
 & COMB-A & 99.9 & 99.8 & 65.9 & 35.1 & 75.2 & 48.0 & 44.0 & 46.0 & 64.2 \\
 & COMB-R & 98.6 & 98.5 & 67.1 & 72.2 & 84.1 & 37.0 & 43.1 & 40.0 & 67.6 \\
 & COMB & 99.9 & 99.8 & \textbf{76.8} & 71.9 & \textbf{87.1} & 45.6 & 45.4 & 45.5 & \textbf{71.5} \\
\cline{1-11}
\multirow[t]{9}{*}{Llama-3.1-8B} & GEN & 99.6 & 99.3 & 54.3 & 61.6 & 78.7 & 46.3 & 59.0 & 52.7 & 68.9 \\
 & ICR & \textbf{100.0} & 89.6 & 61.7 & 60.6 & 78.0 & 67.4 & 83.8 & 75.6 & 77.1 \\
 & AT2 & 99.9 & 99.9 & 72.7 & 57.6 & 82.5 & \textbf{67.9} & 84.9 & \textbf{76.4} & 80.2 \\
 & QR & 99.9 & 99.9 & 71.1 & 78.2 & 87.3 & 67.7 & \textbf{85.0} & 76.3 & 83.2 \\
 & BM25 & 98.7 & 70.2 & 49.6 & 75.6 & 73.5 & 52.7 & 81.6 & 67.2 & 71.1 \\
 & DRAG & 99.6 & \textbf{100.0} & 71.8 & 68.6 & 85.0 & 62.5 & 80.9 & 71.7 & 80.0 \\
 & COMB-A & 99.9 & 100.0 & 76.8 & 79.8 & 89.1 & 63.5 & 84.5 & 74.0 & 83.5 \\
 & COMB-R & 100.0 & 100.0 & 77.2 & \textbf{86.0} & 90.8 & 59.1 & 81.7 & 70.4 & 83.2 \\
 & COMB & 99.9 & 100.0 & \textbf{80.9} & 86.0 & \textbf{91.7} & 65.1 & 84.6 & 74.8 & \textbf{85.4} \\
\cline{1-11}
\multirow[t]{9}{*}{Llama-3.3-70B} & GEN & 98.2 & 98.8 & 49.1 & 61.4 & 76.9 & 46.7 & 43.1 & 44.9 & 64.9 \\
 & ICR & 99.9 & 36.1 & 56.9 & 52.7 & 61.4 & 71.0 & 76.8 & 73.9 & 66.1 \\
 & AT2 & 99.8 & 98.2 & 66.4 & 60.1 & 81.1 & 70.1 & 83.0 & 76.5 & 79.4 \\
 & QR & \textbf{100.0} & 95.4 & 67.7 & 76.3 & 84.8 & \textbf{75.7} & 86.7 & \textbf{81.2} & 83.5 \\
 & BM25 & 99.2 & 69.9 & 49.0 & 77.0 & 73.8 & 61.5 & 83.9 & 72.7 & 73.4 \\
 & DRAG & 99.6 & \textbf{100.0} & 71.9 & 69.3 & 85.2 & 65.1 & 83.1 & 74.1 & 81.0 \\
 & COMB-A & 100.0 & 99.8 & 73.3 & 79.5 & 88.1 & 71.0 & 86.7 & 78.9 & 84.7 \\
 & COMB-R & 100.0 & 99.9 & 75.4 & 83.8 & 89.8 & 57.2 & 85.4 & 71.3 & 82.9 \\
 & COMB & 100.0 & 100.0 & \textbf{79.1} & \textbf{84.4} & \textbf{90.9} & 73.2 & \textbf{87.4} & 80.3 & \textbf{86.9} \\
\cline{1-11}
\multirow[t]{9}{*}{Qwen3-1.7B} & GEN & 94.5 & 96.1 & 40.9 & 53.0 & 71.1 & 54.0 & 24.1 & 39.1 & 59.1 \\
 & ICR & 99.7 & 46.1 & 56.9 & 39.8 & 60.6 & 55.2 & 53.0 & 54.1 & 58.2 \\
 & AT2 & 99.8 & 70.0 & 64.8 & 43.4 & 69.5 & 59.2 & 60.3 & 59.7 & 65.8 \\
 & QR & 99.8 & 69.1 & 65.4 & 43.6 & 69.5 & 61.0 & \textbf{61.7} & \textbf{61.3} & 66.4 \\
 & BM25 & 98.6 & 70.0 & 50.6 & 73.6 & 73.2 & 49.2 & 51.6 & 50.4 & 64.7 \\
 & DRAG & 99.6 & \textbf{100.0} & 74.4 & 70.8 & 86.2 & 53.7 & 55.9 & 54.8 & 74.4 \\
 & COMB-A & 99.8 & 98.4 & 68.6 & 66.5 & 83.3 & \textbf{61.3} & 60.2 & 60.8 & 74.9 \\
 & COMB-R & 99.8 & 99.9 & 73.8 & \textbf{79.8} & \textbf{88.3} & 55.9 & 53.5 & 54.7 & 75.7 \\
 & COMB & \textbf{99.9} & 99.7 & \textbf{75.2} & 75.0 & 87.4 & 61.3 & 60.2 & 60.8 & \textbf{77.4} \\
\cline{1-11}
\multirow[t]{9}{*}{Qwen3-8B} & GEN & 98.6 & 99.8 & 68.0 & 80.8 & 86.8 & 62.6 & 43.6 & 53.1 & 74.2 \\
 & ICR & 99.8 & 38.0 & 55.4 & 42.5 & 58.9 & 59.1 & 50.9 & 55.0 & 57.4 \\
 & AT2 & \textbf{100.0} & 94.4 & 73.2 & 52.2 & 79.9 & 69.4 & 66.4 & 67.9 & 75.4 \\
 & QR & 99.9 & 44.1 & 70.5 & 51.8 & 66.6 & 69.4 & \textbf{67.2} & \textbf{68.3} & 67.2 \\
 & BM25 & 98.7 & 69.5 & 50.4 & 71.8 & 72.6 & 57.5 & 55.0 & 56.2 & 66.5 \\
 & DRAG & 99.6 & \textbf{100.0} & 74.5 & 69.4 & 85.9 & 63.3 & 59.6 & 61.5 & 76.7 \\
 & COMB-A & 100.0 & 99.9 & 80.0 & 82.5 & 90.6 & \textbf{69.9} & 63.6 & 66.7 & 81.6 \\
 & COMB-R & 99.9 & 100.0 & 82.4 & \textbf{90.4} & \textbf{93.2} & 67.2 & 60.2 & 63.7 & 82.1 \\
 & COMB & 100.0 & 100.0 & \textbf{83.9} & 87.2 & 92.8 & 69.9 & 64.2 & 67.0 & \textbf{83.1} \\
\cline{1-11}
\multirow[t]{9}{*}{Qwen3-32B} & GEN & 99.8 & 99.9 & 76.1 & 81.4 & 89.3 & \textbf{63.5} & 48.3 & 55.9 & 76.8 \\
 & ICR & 99.0 & 15.0 & 48.0 & 41.4 & 50.9 & 39.9 & 31.2 & 35.6 & 45.1 \\
 & AT2 & \textbf{100.0} & 79.6 & 67.2 & 51.5 & 74.6 & 61.5 & 58.8 & 60.1 & 69.2 \\
 & QR & 100.0 & 61.6 & 70.5 & 65.9 & 74.5 & 62.6 & \textbf{60.8} & \textbf{61.7} & 69.7 \\
 & BM25 & 99.1 & 69.5 & 50.5 & 71.8 & 72.7 & 53.8 & 48.2 & 51.0 & 64.6 \\
 & DRAG & 99.5 & \textbf{100.0} & 73.8 & 68.7 & 85.5 & 58.2 & 51.7 & 54.9 & 74.0 \\
 & COMB-A & 100.0 & 99.9 & 83.7 & 86.4 & 92.5 & 63.1 & 57.3 & 60.2 & 80.4 \\
 & COMB-R & 99.9 & 99.9 & 84.6 & \textbf{90.8} & \textbf{93.8} & 60.6 & 54.2 & 57.4 & 80.2 \\
 & COMB & 100.0 & 99.9 & \textbf{85.4} & 89.8 & 93.8 & 63.1 & 57.8 & 60.5 & \textbf{81.3} \\
\bottomrule
\end{tabular}
    \caption{Numerical values for Fig.~\ref{fig:attention_results}. Metrics: \benchmarkname{}: \reckf{}, Transfer: Attributability.}
    \label{tab:citention_all_results}
\end{table*}

\begin{figure*}[!htb]
    \includegraphics[]{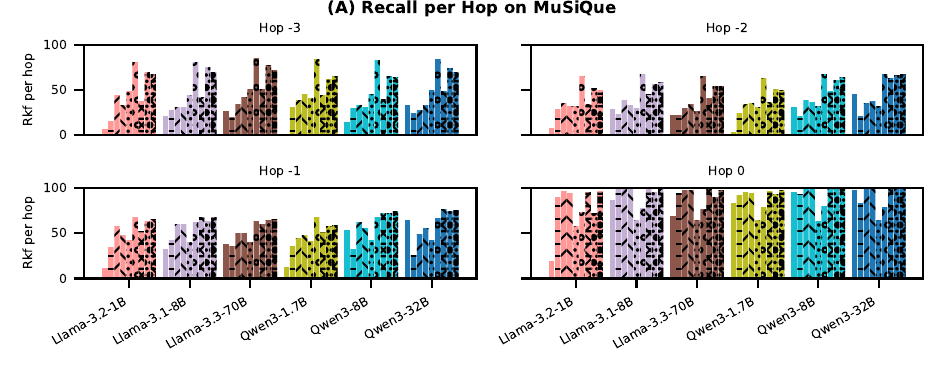}
    \includegraphics[]{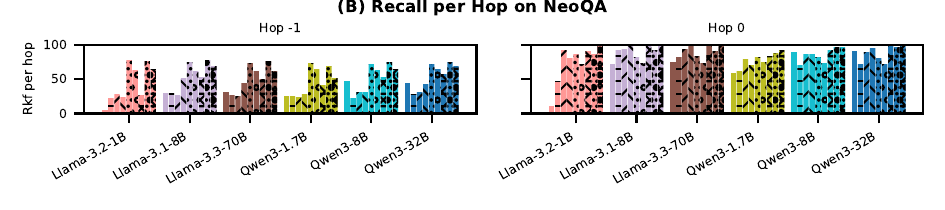}
    \caption{Recall per hop on \musique{} and \neoqa{} \concatenation{} instances for models and citation approaches from §\ref{sec:attention/results}.}
\label{fig:attention_recall_per_hop_neoqa}
\end{figure*}

\end{document}